\documentclass{bmvc2k}

\usepackage{wrapfig}

\title{Functionally Modular and Interpretable Temporal Filtering for Robust Segmentation}

\addauthor{J\"org Wagner\textsuperscript{\scriptsize{1,}}}{Joerg.Wagner3@de.bosch.com}{2}
\addauthor{Volker Fischer}{Volker.Fischer@de.bosch.com}{1}
\addauthor{Michael Herman}{Michael.Herman@de.bosch.com}{1}
\addauthor{Sven Behnke}{behnke@cs.uni-bonn.de}{2}

\addinstitution{
 Bosch Center for Artificial Intelligence,\\
 71272 Renningen, Germany
}
\addinstitution{
 University of Bonn,\\
 Computer Science Institute VI,\\
 Autonomous Intelligent Systems,\\
 Endenicher Allee 19 A,\\
 53115 Bonn, Germany
}

\runninghead{Wagner ET AL.}{Modular and Interpretable Temporal Filtering}

\def\eg{\emph{e.g}\bmvaOneDot}

\newcommand{\norm}[1]{\left\lVert#1\right\rVert}

\begin{document}
\maketitle

\begin{abstract}
The performance of autonomous systems heavily relies on their ability to generate a robust representation of the environment. Deep neural networks have greatly improved vision-based perception systems but still fail in challenging situations, \eg sensor outages or heavy weather. These failures are often introduced by data-inherent perturbations, which significantly reduce the information provided to the perception system. We propose a functionally modularized temporal filter, which stabilizes an abstract feature representation of a single-frame segmentation model using information of previous time steps. Our filter module splits the filter task into multiple less complex and more inter\-pretable subtasks. The basic structure of the filter is inspired by a Bayes estimator consisting of a prediction and an update step. To make the prediction more transparent, we implement it using a geometric projection and estimate its parameters. This additionally enables the decomposition of the filter task into static representation filtering and low-dimensional motion filtering. Our model can cope with missing frames and is trainable in an end-to-end fashion. Using photorealistic, synthetic video data, we show the ability of the proposed architecture to overcome data-inherent perturbations. The experiments especially highlight advantages introduced by an interpretable and explicit filter module.
\end{abstract}

\section{Introduction}
\label{sec:intro}
The performance of autonomous systems, such as mobile robots or self-driving cars, is heavily influenced by their ability to generate a robust representation of the current environment. Errors in the environment representation are propagated to subsequent processing steps and are hard to recover. For example, a common error is a missed detection of an object, which might lead to a fatal crash. In order to increase the reliability and safety of autonomous systems, robust methods for observing and interpreting the environment are required. 

Deep learning based methods have greatly advanced the state-of-the-art of perception systems. Especially vision-based perception benchmarks (\eg Cityscapes~\cite{cordts_2016_CVPR} or Caltech~\cite{dollar_2019_CVPR}) are dominated by approaches utilizing deep neural networks. From a safety perspective, a major disadvantage of such benchmarks is that they are recorded at daytime under idealized environment conditions. To deploy autonomous systems in an open world scenario without any human supervision, one not only has to guarantee their reliability in good conditions, but also has to make sure that they still work in challenging situations (\eg sensor outages or heavy weather). One source of such challenges are perturbations inherent in the data, which significantly reduce the information provided to the perception system. We denote failures originating from data-inherent perturbations in accordance to the classification of uncertainties~\cite{Kiureghian_2019_Safety, kendall_2017_arxiv} as aleatoric failures. These failures cannot be resolved using a more powerful model or additional training data. To solve aleatoric failures, one has to enhance the information provided to the perception system. This can be achieved by fusing the information of multiple sensors, utilizing context information or by considering temporal information. A second class of failures are epistemic failures, which are model or dataset dependent. They can be mitigated by using more training data and/or a more powerful model~\cite{Kiureghian_2019_Safety}. 

\begin{figure}
	\includegraphics[width=\linewidth]{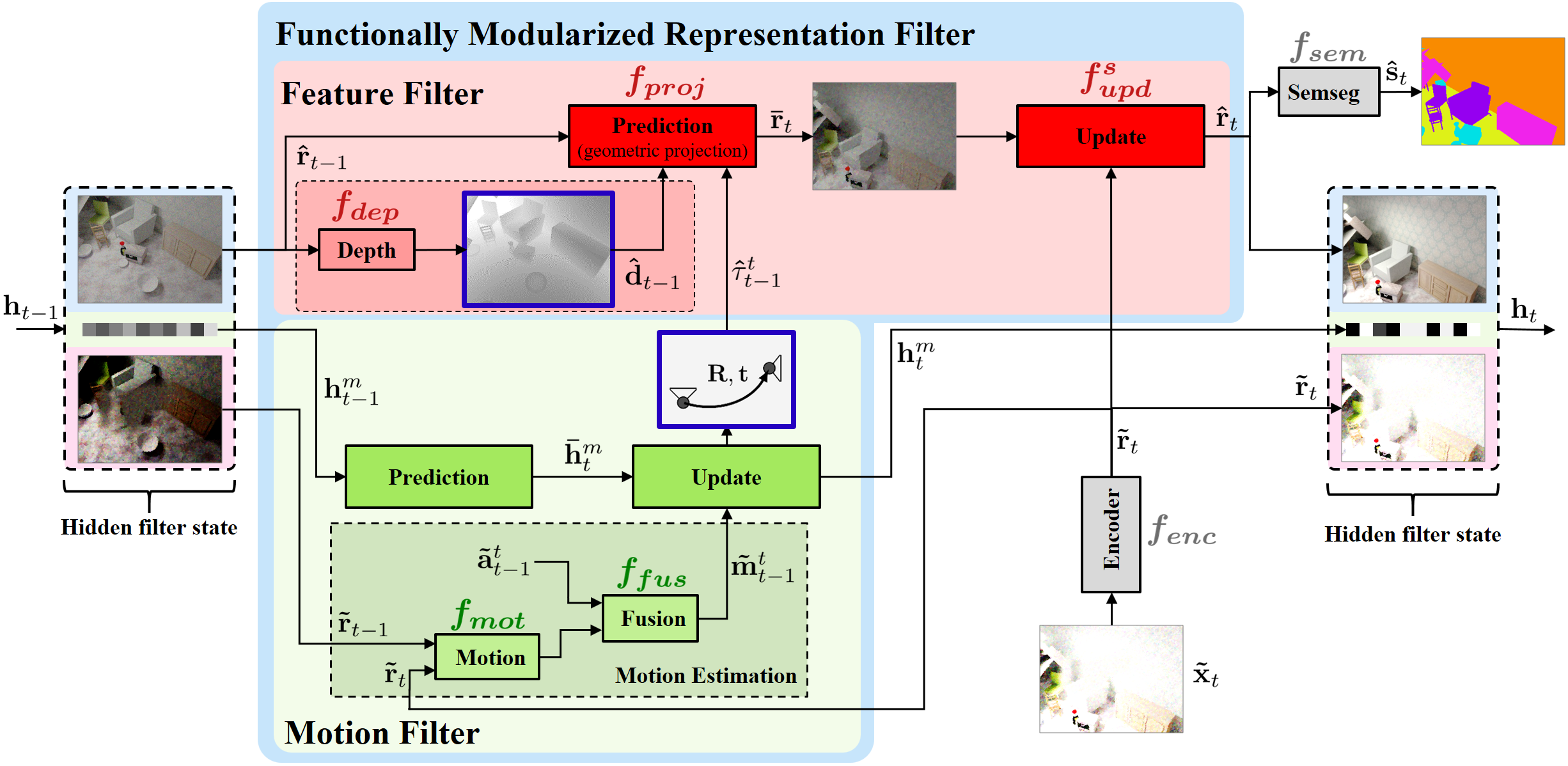}
	\vspace{-0.71cm}
	\caption{Overview of the functionally modularized representation filter with its subcomponents. For visualization purposes we use images to represent feature maps. }
	\label{fig:filter_overview}
	\vspace{-0.53cm}
\end{figure}

In this work, we focus on tackling aleatoric failures of a single frame semantic segmentation model using temporal consistency. Temporal integration is achieved by recurrently filtering a representation of the model using a functionally modularized filter (Fig.~\ref{fig:filter_overview}). In contrast to other available approaches, our filter consists of multiple submodules, decomposing the filter task into less complex and more transparent subtasks. The basic structure of the filter is inspired by a Bayes  estimator, consisting of a prediction step and an update step. 
We model the prediction of the representation as an explicit geometric projection given estimates of the scene geometry and the scene dynamics. The scene geometry and dynamics are represented as a per pixel depth and a 6-DoF camera motion. Both parameters are estimated within the filter using two task-specific subnetworks.

The decomposition of the prediction task into a model-based transformation as well as a depth and a motion estimation introduces several advantages. Instead of having to learn dynamics of a high-dimensional representation, we now can model motion separately in a low-dimensional space. The overall filter can therefore be subdivided into two subfilters: A motion filter, which predicts and integrates low-dimensional camera motion and a feature filter, which handles the integration and prediction of abstract scene features. 
 
An advantage of our approach is its improved transparency, interpretability and explicitness. Within the filter, we estimate two human interpretable representations: a depth map and a camera motion (Fig~\ref{fig:filter_overview}, blue boxes). These representations can be used to inspect the functionality of the model, to split the filter into pre-trainable subnetworks, or to debug and validate network behavior. Besides its modularity, our model is trainable in an end-to-end fashion. 
In contrast to other methods, the proposed filter also works in cases when the current image is not available. Methods that for example use the optical flow fail in such situations due to their inability to compute a meaningful warping.

\section{Related Work}
\label{sec:related_work}
In this section, we give an overview of approaches that use temporal information to improve segmentation models against aleatoric failures. \\

\noindent\textbf{Feature-level temporal filtering.} A common approach to temporally stabilize network predictions are feature-level filters. These filters are applied to one or several feature representations, which are integrated using information of previous time steps. Several works implement such a filter using fully learned, model-free architectures. \citet{fayyaz_2016_arxiv} and \citet{valipour_2017_WACV} generate a feature representation for each image in a sequence and use recurrent neural networks to temporally filter them. \citet{jin_2016_arxiv} utilize a sequence of previous images to predict a feature representation of the current image. The predicted representation is fused with the one of the current image and propagated through a decoder network. The \textit{Recurrent Fully Convolutional DenseNet}~\cite{wagner_2018_ESANN} utilizes a hierarchical filter concept to increase the robustness of a segmentation model. Being model-free, these filters require many parameters and are therefore harder to train. Due to their low interpretability, it is quite difficult to include constraints and to inspect or debug their behavior.

A second class of feature-level filters utilizes a partially model-based approach to integrate features. These approaches use an explicit model to implement the temporal propagation of features and learn a subnetwork to fuse the propagated features with features of the current time step. A common model to implement the propagation is optical flow. The replacement field parametrizing the flow can be predicted in the model \cite{vu_2018_arxiv} or computed using classical methods~\cite{gadde_2017_ICCV, Nilsson_2016_CoRR}. These models are well suited to reduce epistemic failures, but often fail to resolve aleatoric failures. This is due to their dependence on the availability of the current frame. More sophisticated feature propagation models exist~\cite{zhou_2017_CVPR, mahjourian_2016_arxiv, mahjourian_2018_arxiv, yin_2018_arxiv}, which additionally constrain the transformation. Such a model was recently used to temporally aggregate learned features within a multi-task model \cite{radwan_2018_arxiv}. Our model is also partially model-based and utilizes a more sophisticated propagation model similar to~\citet{radwan_2018_arxiv}. In contrast to all presented model-based approaches, our filter is not dependent on the availability of the current frame. \\

\noindent\textbf{Post-processing based temporal integration.} Some approaches use post-processing steps to integrate the predictions of single-frame segmentation models. \citet{lei_2016_ECCV} propose the \textit{Recurrent-Temporal Deep Field} model for video segmentation, which combines a convolutional neural network, a recurrent temporal restricted Boltzmann machine, as well as a conditional random field. \citet{kundu_2016_CVPR} propose a long-range spatio-temporal regularization using a conditional random field operating on a feature space, optimized to minimize the distance between features associated with corresponding points. 

Our temporal integration approach differs from these post-processing methods, due to the integration of rich feature representations instead of segmentations. The modular structure of our filter, with its human interpretable representations, makes it also more transparent. \\

\noindent\textbf{Spatio-temporal fusion.} Other approaches build semantic video segmentation networks using spatio-temporal features. \citet{tran_2016_CVPR} and \citet{zhang_2014_ICVRV} use 3D convolutions to compute such features. The \textit{Recurrent Convolutional Neural Network} of \citet{pavel_2015_IJCNN} is another spatio-temporal architecture. This method uses layer-wise recurrent self-connections as well as top-down connections to stabilize representations. These approaches require a large number of parameters. Additionally, it is quite difficult to integrate physical constraints.

\section{Functionally Modularized Temporal Filtering}
\label{sec:temp_filtering}
The aim of this work is to improve the robustness of a deep neural network $f$, which receives a measurement $\mathbf{\tilde{x}}_{t}$ and produces a pixel-wise semantic segmentation~$\mathbf{\tilde{s}}_{t}$. We assume the model consists of two parts: a feature encoder $f_{enc}$ and a semantic decoder $f_{sem}$. The feature encoder generates an abstract feature representation $\mathbf{\tilde{r}}_{t}$ of the image $\mathbf{\tilde{x}}_{t}$. This representation is up-sampled and refined by the semantic decoder $f_{sem}$ to produce a dense segmentation $\mathbf{\tilde{s}}_{t}$:
\begin{equation}
\mathbf{\tilde{s}}_{t} = f(\mathbf{\tilde{x}}_{t};\theta) = f_{sem}(\mathbf{\tilde{r}}_{t};\theta_{sem}) = f_{sem}( f_{enc}(\mathbf{\tilde{x}}_{t};\theta_{enc});\theta_{sem}). 
\label{eq:semseg_model}
\end{equation}

Due to data-inherent perturbations, the representation $\mathbf{\tilde{r}}_{t}$ is an approximation of the true feature representation without perturbations. Using a temporal filter $f_{FM}$, we try to improve the estimate of the features $\mathbf{\hat{r}}_{t}$ and, as a result, the estimate of the semantic decoder:
\begin{equation}
\mathbf{\hat{s}}_{t}=f_{sem}(\mathbf{\hat{r}}_{t};\theta_{sem}) = f_{sem}( f_{FM}(\mathbf{\tilde{r}}_{t}, \mathbf{h}_{t-1} ;\theta_{FM});\theta_{sem}).
\label{eq:general_filter}
\end{equation}
All prior knowledge about scene features and dynamics, aggregated from previous time steps, is encoded in the hidden state $\mathbf{h}_{t-1}$.

Framing Eq.~\ref{eq:general_filter} in the context of a Bayesian estimator, the recurrent filter module has to propagate the belief about the hidden state one time step into the future, update the belief using the current filter input $\mathbf{\tilde{r}}_{t}$, and compute an improved estimate $\mathbf{\hat{r}}_{t}$ of the true feature representation. To make our filter module more transparent, we adopt the basic structure of a Bayesian estimator and split the filter into a prediction $ f_{pred}$ and update $f_{upd}$ module:
\begin{equation}
	\mathbf{\hat{r}}_{t}=f_{FM}(\mathbf{\tilde{r}}_{t},\mathbf{h}_{t-1};\theta_{FM}) = f_{upd}(\mathbf{\tilde{r}}_{t}, f_{pred}(\mathbf{\tilde{r}}_{t}, \mathbf{h}_{t-1} ;\theta_{pred});\theta_{upd}).
\label{eq:bayesian_filter}
\end{equation}
The prediction module propagates the hidden state, while the update module refines it using information in $\mathbf{\tilde{r}}_{t}$ for deriving an improved estimate of the encoder representation $\mathbf{\hat{r}}_{t}$. The prediction module therefore has to learn the complex dynamics of a high-dimensional hidden-state. To increase explainability and divide the prediction task into easier subtasks, we split the hidden state into a high-dimensional static state $\mathbf{h}_{t-1}^{s}=(\mathbf{\hat{r}}_{t-1}, \mathbf{\tilde{r}}_{t-1})$ encoding all scene features and a low-dimensional dynamic state $\mathbf{h}_{t-1}^{m}$ encoding scene dynamics (Fig.~\ref{fig:filter_overview}). 

The prediction of $\mathbf{h}_{t-1}^{s}$ can now be performed fully model-based using a geometric projection $f_{proj}$. This is possible, since the prediction only has to account for spatial feature displacements. To compute a valid projection, estimates of the scene geometry and the scene dynamics are required. We encode the scene geometry as a per-pixel depth and derive it from the static hidden state by means of a depth decoder $f_{dep}$. A 3D rigid transformation $\tau_{t-1}^{t}$ is used to characterize the scene dynamics, assuming the dynamics are dominated by camera motion. The predicted static hidden state is updated in a second module $f_{upd}^{s}$ using the new information of the input $\mathbf{\tilde{r}}_{t}$. The two modules $f_{proj}$ and $f_{upd}^{s}$ form the static feature filter. 

Scene dynamics, represented by a low-dimensional state $\mathbf{h}_{t-1}^{m}$, are filtered in a second subfilter. A motion estimation module is used to project from the high-dimensional scene feature space into a low-dimensional motion feature space. The transformation is fully learned, enabling the model to generate a representation well-suited for motion integration.

By decoupling motion and scene features, it is much easier to incorporate auxiliary information such as acceleration data $\mathbf{\tilde{a}}_{t-1}^{t}$ of the sensor. This kind of information can now be fused (see module $f_{fus}$ in Fig.~\ref{fig:filter_overview}) much more targeted with the appropriate motion features derived from image pairs (see module $f_{mot}$ in Fig.~\ref{fig:filter_overview}). An additional advantage of the decoupling is a global modelling of camera motion. The motion is guaranteed to be consistent across spatial scene features and can be estimated using correlations across full image pairs. 

Another way of looking at our model is that it consists of an undelying multi-task model:
\begin{equation}
	f_{MT}(\cdot;\theta_{MT})=f_{enc}(\cdot;\theta_{enc}) \cup f_{sem}(\cdot;\theta_{sem}) \cup f_{dep}(\cdot;\theta_{dep}) \cup f_{mot}(\cdot;\theta_{mot}),
	\label{eq:multi_task_model}
\end{equation} 
predicting a segmentation, a depth map and a 3D rigid transformation. The encoder representation is integrated over time using an additional filter module, which utilizes  decoder outputs to propagate previous knowledge. As a result, decoders operate on a filtered encoder representation or are filtered separately (see $f_{mot}$), making the functionality of our model not dependent on new inputs $\mathbf{\tilde{r}}_{t}$. This property sets our filter apart from other approaches. 

The overall filter is set-up to increase transparency and interpretability, by modularizing functionalities, using model-based computations, and introducing human interpretable representations. Compared to other architectures, it is hence much easier to debug and validate the model, inspect intermediate results and pre-train subnetworks. These properties are also particularly relevant with regard to safety analysis. From a multi-task perspective, the two auxiliary tasks may also benefit segmentation, due to the implicit regularization~\cite{ruder_2017_arxiv}.

\subsection{Feature Filter}
\label{sec:feature_filter}
\noindent\textbf{Depth Estimation.} We compute a per-pixel depth $\mathbf{\hat{d}}_{t-1}$ using a decoder network operating on the filtered representation $\mathbf{\hat{r}}_{t-1}$. The depth decoder consists of three convolutional layers with kernel size 3$\times$3, 1$\times$1, and 1$\times$1, respectively. We apply batch normalization and use ReLU nonlinearities in each layer. The predicted depth is therefore always positive and valid. The number of features in the first two layers is set to $384$ and the last layer predicts one value per pixel. Instead of directly predicting depth values, we let the decoder provide the inverse depth $\mathbf{\hat{z}}_{t-1}=1/\mathbf{\hat{d}}_{t-1}$, which puts less focus on wrong predictions in larger distance. For supervision during training, we use two losses. A L1 loss on the inverse depth:
\begin{equation}
\mathcal{L}_{depth}^{L1} =  \sum_{i,j} |\mathbf{z}_{t-1}(i,j) - \mathbf{\hat{z}}_{t-1}(i,j)|,
\label{eq:l1_depth_loss}
\end{equation}
 and a scale-invariant gradient loss~\cite{ummenhofer_2017_CVPR} to take dependencies of depths into account:
\begin{align}
	\mathbf{g}_{h}[n](i,j) &= \left(  \frac{n(i+h,j)-n(i,j)}{|n(i+h,j)|+|n(i,j)|} , \frac{n(i,j+h)-n(i,j)}{|n(i,j+h)|+|n(i,j)|}  \right)^{\top} ,\\
	\mathcal{L}_{depth}^{sig} &= \sum_{h\in\{1,2,4\}} \sum_{i,j} \norm{ \mathbf{g}_{h}[\mathbf{z}_{t-1}](i,j) - \mathbf{g}_{h}[\mathbf{\hat{z}}_{t-1}](i,j) }_{2}.
\label{eq:gradient_depth_loss}
\end{align}

\noindent\textbf{Prediction / Geometric Projection.} To make the prediction of features more explicit, we use a geometric projection~\cite{zhou_2017_CVPR, mahjourian_2016_arxiv}. Let $\mathbf{p}_{t-1}$ be the coordinates of each pixel at time step $t-1$ and $K$ the camera intrinsic matrix. The projection can be implemented as:
\begin{equation}
	\mathbf{p}_{t}^{(i,j)} \sim K \mathbf{\hat{\tau}}_{t-1}^{t} \mathbf{\hat{d}}_{t-1}^{(i,j)} K^{-1} \mathbf{p}_{t-1}^{(i,j)}.
\label{eq:geo_proj}
\end{equation}
To keep the notation short, we avoided all conversions related to homogeneous coordinates. The coordinates $\mathbf{p}_{t}^{(i,j)}$ are continuous and have to be discretized. Additionally, it is necessary to account for ambiguities, in cases where multiple pixels at time step $t-1$ are assigned to the same pixel at time step $t$. We resolve these ambiguities by using the transformed pixels with smaller depth (objects closer to the sensor). This projection is differentiable with respect to scene features. In contrast to other methods, our implementation does not depend on information of time step $t$. This is an important property for resolving aleatoric failures. \\

\noindent\textbf{Update / Feature Fusion.} The update module enables the network to weight the predicted representation $\mathbf{\bar{r}}_{t}$ and the input representation $\mathbf{\tilde{r}}_{t}$ depending on the information of these two representations (data-dependent weighting). For each pixel position a weighting value $\mathbf{i}_{t}^{(i,j)}$ is estimated that indicates whether one can rely on prior knowledge ($\mathbf{\bar{r}}_{t}$) or on information of new inputs ($\mathbf{\tilde{r}}_{t}$). This weight matrix $\mathbf{i}_{t}$ is calculated similarly to convolution LSTM gates, but contains only one value per pixel instead of one value per pixel and feature:
\begin{equation}
\mathbf{i}_{t} = \sigma(\mathbf{W}_{hid} \ast \mathbf{\bar{r}}_{t} + \mathbf{W}_{in} \ast \mathbf{\tilde{r}}_{t} + \mathbf{b}_{i} ).
\label{eq:feature_update_gate}
\end{equation}
The convolutional operator is indicated by $\ast$, $\mathbf{W}_{hid}$ and $\mathbf{W}_{in}$ are 3$\times$3 kernels, and $\mathbf{b}_{i}$ is a bias. Using $\mathbf{i}_{t}$ and element-wise multiplications $\circ$, the update module computes:
\begin{equation}
	\mathbf{\hat{r}}_{t} =  (1 - \mathbf{i}_{t}) \circ \mathbf{\bar{r}}_{t} +  \mathbf{i}_{t} \circ \mathbf{\tilde{r}}_{t}.
	\label{eq:feature_update}
\end{equation}

\subsection{Motion Filter}
\label{sec:motion_filter}
The motion filter consists of two components: a motion estimator and a motion integration module. Both modules are model-free and learned during training (see Section~\ref{sec:implementation_details}). \\

\begin{figure}[h]
	\vspace{-0.0em}
	\includegraphics[width=\linewidth]{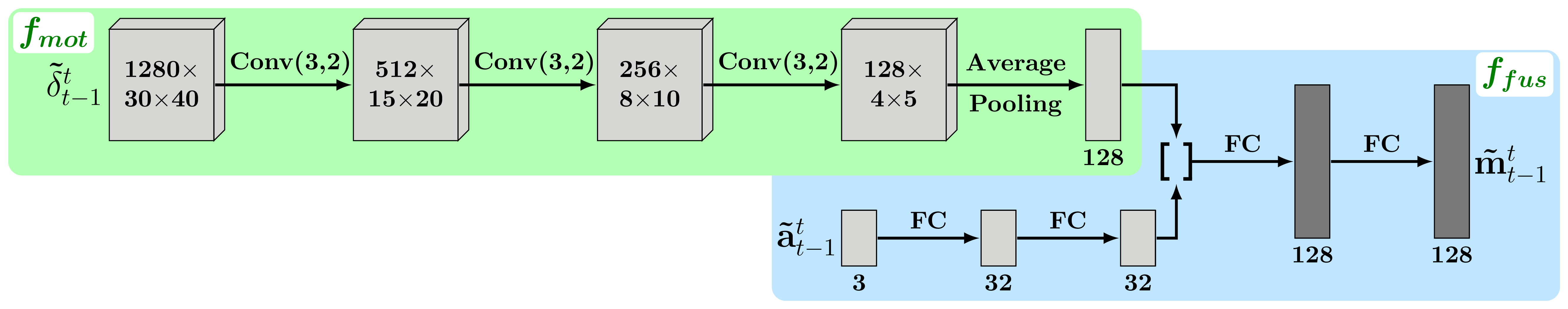}
	\vspace{-1.8em}
	\caption{Motion estimation module. Conv(f,s): convolutional layer with f filters and a stride of s; FC: fully connected layer; $\mathbf{[\hspace{0.15cm}]}$: Concatenation of features.}
	\label{fig:motion_estimation}
	\vspace{-1.0em}
\end{figure}

\noindent\textbf{Motion Estimation.} Using the decoder network $f_{mot}$, the motion estimation module learns a projection from the high-dimensional scene feature space to the low-dimensional motion feature space. To stabilize motion estimates, the projected features are combined in the fusion module $f_{fus}$ with acceleration data of the camera. The motion estimation module is depicted in Fig.~\ref{fig:motion_estimation}. Pairs of encoder representations $\mathbf{\tilde{\delta}}_{t-1}^{t} = [\mathbf{\tilde{r}}_{t-1},\mathbf{\tilde{r}}_{t}]$ concatenated along the feature dimension are used as the input of the motion decoder. We apply batch normalization and utilize ReLU nonlinearities in each convolutional and fully connected layer.
\\

\noindent\textbf{Temporal Motion Integration.} If the input is noisy, the features $\mathbf{\tilde{m}}_{t-1}^{t}$ computed in the motion estimator contain only limited information. In order to still obtain a meaningful motion estimate, we integrate motion features over time in a model-free filter. This filter is based on a gated recurrent unit (GRU)~\cite{cho_2014_arxiv} and defined by:
\begin{align}
	\mathbf{k}^{t}_{t-1} &= \sigma( \mathbf{W}_{\tilde{m}k}  \mathbf{\tilde{m}}_{t-1}^{t} + \mathbf{W}_{hk}  \mathbf{h}_{t-1}^{m} + \mathbf{b}_{k}), \forall \mathbf{k} \in \left\{\mathbf{o},\mathbf{u}\right\}, \\
	\mathbf{c}^{t}_{t-1} &= \sigma( \mathbf{W}_{\tilde{m}c}  \mathbf{\tilde{m}}_{t-1}^{t} + \mathbf{o}_{t-1}^{t} \circ ( \mathbf{W}_{hc}  \mathbf{h}_{t-1}^{m}) + \mathbf{b}_{c}) , \\
	\mathbf{h}_{t}^{m} &= (1 - \mathbf{u}^{t}_{t-1}) \circ \mathbf{h}_{t-1}^{m} + \mathbf{u}^{t}_{t-1} \circ \mathbf{c}^{t}_{t-1}.
	\label{eq:motion_loss}
\end{align}

\noindent To infer the 3D rigid camera transformation $\mathbf{\hat{\tau}}_{t-1}^{t}=\{\mathbf{\hat{T}}_{t-1}^{t},\mathbf{\hat{R}}_{t-1}^{t}\}$ from the filtered hidden state $\mathbf{h}_{t}^{m}$, we propagate it through two additional fully connected layers with $128$ and $6$ features, respectively. The output layer predicts the translation vector $\mathbf{\hat{T}}_{t-1}^{t}$ and the sinus of the rotation angles $(\sin \alpha_{t-1}^{t}, \sin \beta_{t-1}^{t}, \sin \gamma_{t-1}^{t})^{\top}$. The first layer applies batch normalization and uses a ReLU nonlinearity and the output layer uses no nonlinearity for the translation vector and clips the angle sinus estimates to  $[-1,1]$. Based on the clipped values, we compute the rotation matrix $\mathbf{\hat{R}}_{t-1}^{t}$. \\

\noindent\textbf{Motion Supervision.} All parameters of the motion filter are trained using ground-truth camera translation vectors $\mathbf{T}_{t-1}^{t}$ and rotation matrices $\mathbf{R}_{t-1}^{t}$. The losses are based on the relative transformation between the predicted and ground-truth motion as defined by~\cite{Vijayanarasimhan_2017_corr}:
\begin{align}
	\mathcal{L}_{motion}^{trans} &= \Delta T = \norm{ \textrm{inv} \left( \mathbf{\hat{R}}_{t-1}^{t} \right) \left( \mathbf{T}_{t-1}^{t} - \mathbf{\hat{T}}_{t-1}^{t} \right) }_{2}^{2},\label{eq:motion_loss_trans}\\
	\mathcal{L}_{motion}^{rot} &= \Delta R = \arccos\left( \min \left( 1, \max \left( -1, \frac{\textrm{trace}(   \textrm{inv} \left( \mathbf{\hat{R}}_{t-1}^{t} \right) \mathbf{R}_{t-1}^{t}  )-1}{2} \right) \right) \right). \label{eq:motion_loss_rot}
\end{align}

\section{Experiments}
\subsection{Implementation Details}
\label{sec:implementation_details}
\noindent\textbf{Dataset.}
We evaluate our filter using the \textit{SceneNet RGB-D}~\cite{mccormac_2017_ICCV} dataset which consists of 5M photorealistically rendered RGB-D images recorded from 15K indoor trajectories. Besides the camera motion, all scenes are assumed static. Due to its simulated nature, the dataset provides labels for semantic segmentation, depth estimation, and camera motion estimation. We split the training data into a training and validation set and use the provided validation data to setup the test set. For training, we use all non-overlapping sequences of length 7 generated from the training trajectories. The test set is constructed by sampling 5 non-overlapping sequences of length 7 from each test trajectory, resulting in 5,000 test sequences. 

To add aleatoric uncertainty, all sequences are additionally perturbed with noise, clutter, and changes in lighting conditions. Noise is simulated by adding zero-mean Gaussian noise to each pixel. Clutter is introduced by setting subregions of each image to the pixel mean, computed on a per sequence basis. The clutter is generated once per sequence and applied to each frame. Thus, the resulting clutter pattern is the same in each frame, comparable to dirt on the camera lens. To simulate rapid changes in lighting conditions, we increase or decrease the intensity of frames by a random value and let this offset decay over time. Such a noise pattern occurs, for example, when the light is suddenly switched off in a room. We include a more detailed description of the used perturbations in the supplementary material. \\

\begin{figure}[h]
	\centering
	\vspace{-0.17cm}
	\includegraphics[trim={0mm 0mm 0mm 10mm},clip,width=0.96\linewidth]{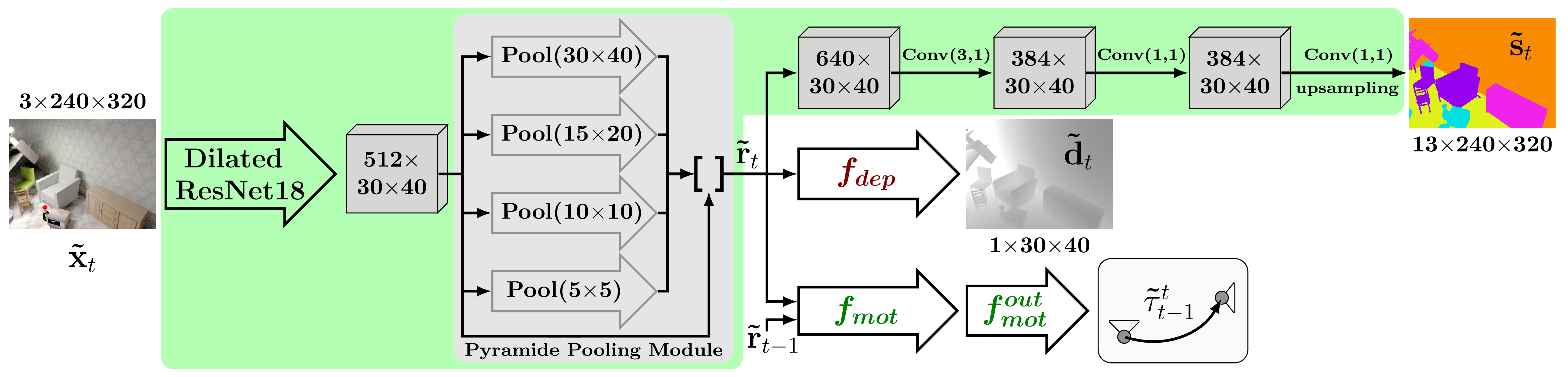}
	\vspace{-0.25cm}
	\caption{Multi-task PSPNet. Conv(f,s): convolutional layer with f filters and a stride of s; Pool(s): Pooling level with kernel size s producing 32 features.}
	\label{fig:MSPSNet}
	\vspace{-0.6em}
\end{figure}

\noindent\textbf{Unfiltered Baseline.}
We use the \textit{Pyramid Scene Parsing Network} (PSPNet)~\cite{zhao_2017_CVPR} as the basis for all architectures~(Fig.~\ref{fig:MSPSNet}, highlighted in green). The used PSPNet is comparatively small to keep the computational effort and the required memory of the filtered models manageable. 

To train our filter module, we additionally need ground-truth depth maps and camera motions. In order to make the comparison of the resulting filtered architecture with the unfiltered baseline fairer, we use a multi-task version of the PSPNet (MPSPNet) in the evaluation. This model operates on image pairs and additionally predicts camera motion and per-pixel depth maps. It can thus also take advantage of all the benefits of multi-task learning~\cite{ruder_2017_arxiv}. The full MPSPNet (see Fig.~\ref{fig:MSPSNet}) uses the depth decoder $f_{dep}$ introduced in Section~\ref{sec:feature_filter} as well as the motion decoder $f_{mot}$ introduced in Section~\ref{sec:motion_filter}. To predict a valid rigid transformation, we reuse the last fully connected layer of the motion filter $f_{mot}^{out}$.

\noindent\textbf{Filtered models.}
Building upon MPSPNet, we set-up our filtered version using the functionally modularized filter concept introduced in Section~\ref{sec:temp_filtering}. We call our filtered model FMTNet.

\begin{wrapfigure}[8]{r}{0.57\textwidth}
	\centering
	\vspace{-0.45cm}	
	\scalebox{0.075}{
		\includegraphics[]{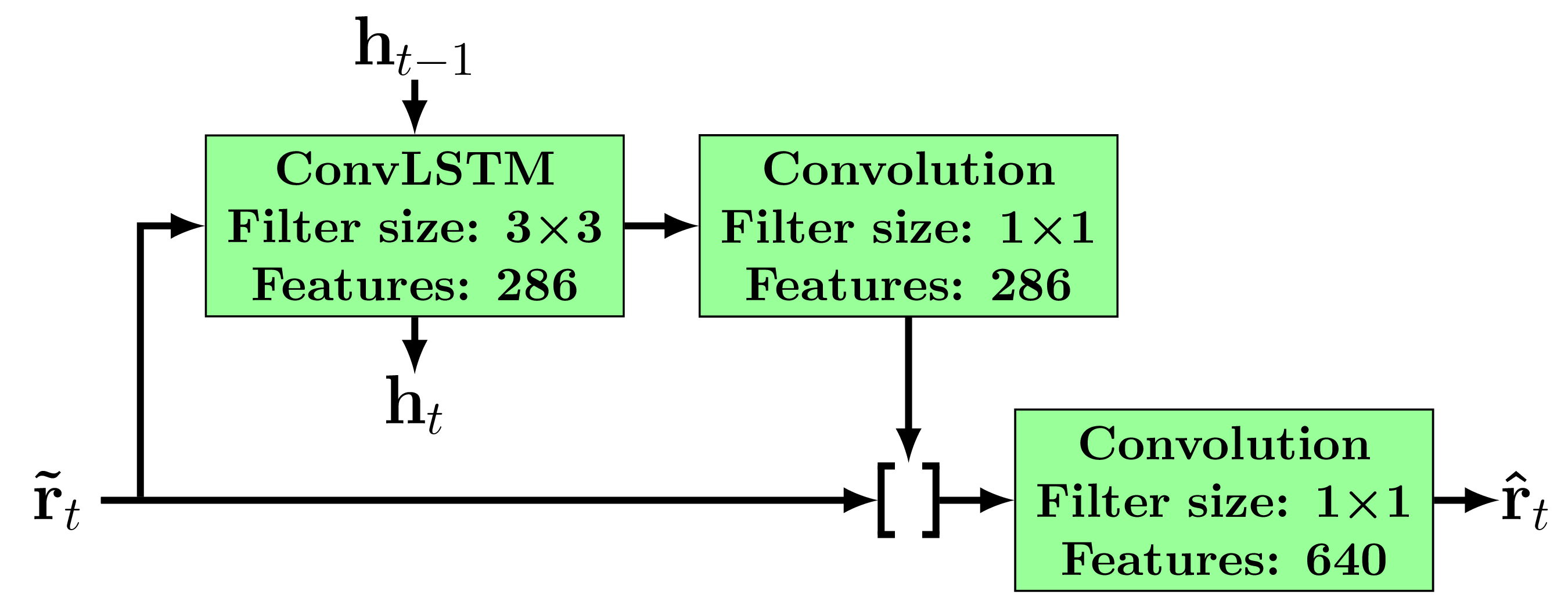}
	}
	\vspace{-0.3cm}	
	\caption{Structure of the filter module MFF.}
	\label{fig:MFF_Filter}	
\end{wrapfigure}

As an additional temporally filtered baseline, we use a model-free, feature-level filter. Such a filter is well suited to solve aleatoric failures, as it does not necessarily require information of the current frame. We use a filter module (denoted by MFF) similar to the one introduced in~\citet{wagner_2018_ESANN} (Fig.~\ref{fig:MFF_Filter}). This filter module receives the representation $\mathbf{\tilde{r}}_{t}$ of MPSPNet as input and generates an improved estimate $\mathbf{\hat{r}}_{t}$ (cf. Eq.~\ref{eq:general_filter}). In the following, 
we will refer to the MPSPNet with model-free filter MFF as MFF-MPSPNet. To be comparable with respect to the filter complexity, the number of parameters in the filter MFF matches the number of parameters in our modularized filter. In the case of our filter, we count the parameters of the depth and motion decoder to the filter, since these decoders are required for filtering. The use of all three decoders in the MFF-MPSPNet guarantees comparable training signals, but is not necessary. Hence, we do not assign the depth and motion decoder weights to the filter MFF, resulting in MFF-MPSPNet having 1.4 times the parameters of FMTNet. \\

\noindent\textbf{Training Procedure.} All models MPSPNet, FMTNet, and MFF-MPSPNet are trained using the multi-task loss introduced by \citet{kendall_2017_arxiv}, which learns the optimal weighting between the cross-entropy segmentation loss, the two depth losses, as well as the two motion losses. We train using Adam~\cite{kingma_2014_arxiv} with a weight decay of 0.0001 and apply dropout with probability 0.1 in the decoders. All components of FMTNet and MFF-MPSPNet that do not belong to the filter are initialized with the corresponding weights of the trained MPSPNet. 

Due to its modularity, we can additionally pre-train two components of our filter. First, we pre-train all weights of the motion filter, while keeping the encoder weights fixed. Second, we pre-train the weights of the feature update module as well as the encoder, while keeping all decoders fixed. The second training is performed with sequences containing the same image, perturbed with aleatoric noise. Finally, we fine-tune the overall architecture.

\subsection{Evaluation}
\label{sec:evaluation}
To evaluate the segmentation performance, we use the Mean Intersection over Union (Mean IoU) on test sequences, computed with 13 classes: bed, books, ceiling, chair, floor, furniture, objects, painting, sofa, table, TV, wall, and window.
In the following two experiments, we first evaluate the motion filter and the update module of the feature filter on toy-like data. In the third experiment, we compare our approach with the unfiltered and filtered baseline using the test dataset described in Section~\ref{sec:implementation_details}. \\

\begin{wrapfigure}[20]{r}{0.5\textwidth}
	\centering
	\vspace{-0.34cm}
	\begin{minipage}[b]{0.5\textwidth}
			\begin{center}
				\scalebox{0.75}{
					\begin{tabular}[]{|c||c|c|c|c|}
						\hline
						 &  Frame 1  		&  Frame 2  	& Frame 3  		& Frame 4  		\\
						\hline 
						Mean IoU 			&  $32.8\,\%$ 	& $42.0\,\%$ 	& $46.2\,\%$ 	& $46.8\,\%$ 	\\	 
						\hline			
					\end{tabular}		
				}		 
			\end{center}
			\vspace{-0.35cm}
			\makeatletter\def\@captype{table}\makeatother 
			\caption{Mean IoU of FMTNet on static toy-data computed for each frame in the sequence.}
			\label{table:static_integration}
	\end{minipage}\vspace{0.1cm}
	\begin{minipage}[b]{0.5\textwidth}
			\centering
			\scalebox{0.045}{
				\includegraphics[]{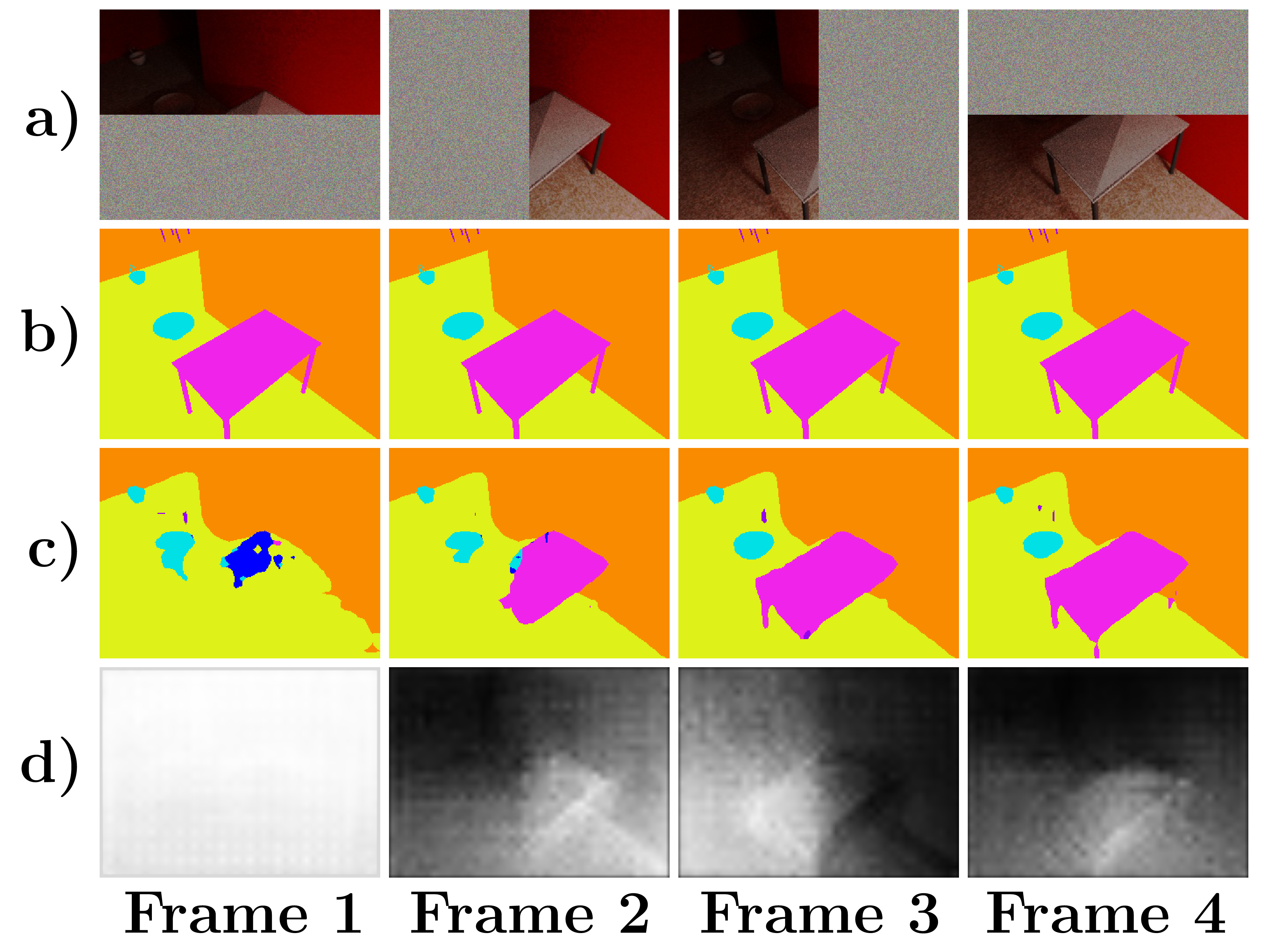}
			}
			\vspace{-0.3cm}	
			\caption{Static toy-data. a): Occluded input sequence; b): Ground-truth semantic segmentation; c): Semantic prediction of FMTNet; d): Update gate $\mathbf{i}_{t}$, white corresponds to a value of one (use new information).}
			\label{fig:static_integration}	
	\end{minipage}
\end{wrapfigure}

\noindent\textbf{Static Feature Integration.} 
To test the functionality of the feature update module, we use a separate static toy-dataset with sequences of length four (see Fig.~\ref{fig:static_integration}\textcolor{red}{a}). Each frame of a
sequence contains the same clean image (random image of the \textit{SceneNet RGB-D} dataset without any of the in Section~\ref{sec:implementation_details} introduced aleatoric perturbations), 50$\%$ of which is replaced by Gaussian noise. We fine-tune the encoder network $f_{enc}$ and feature update module $f_{upd}^{s}$ of the FMTNet on the toy-data. Due to the static nature of the sequences, we remove the motion filter and use the identity transformation (static camera) in the feature filter. 

As shown in Fig.~\ref{fig:static_integration}, FMTNet integrates information over time. It has learned a meaningfull data-dependent weighting between previous information stored in the hidden filter state and information provided by new frames (see weights $\mathbf{i}_{t}$ in Fig.~\ref{fig:static_integration}\textcolor{red}{d}). The same behavior can be seen in Tab.~\ref{table:static_integration}, which reports the Mean IoU on a per-frame basis, computed using 300,000 test sequences. The performance of our model increases over time due to new information. \\

\begin{figure}[h]
		\vspace{-0.1cm}
		\centering
		\includegraphics[width=\linewidth]{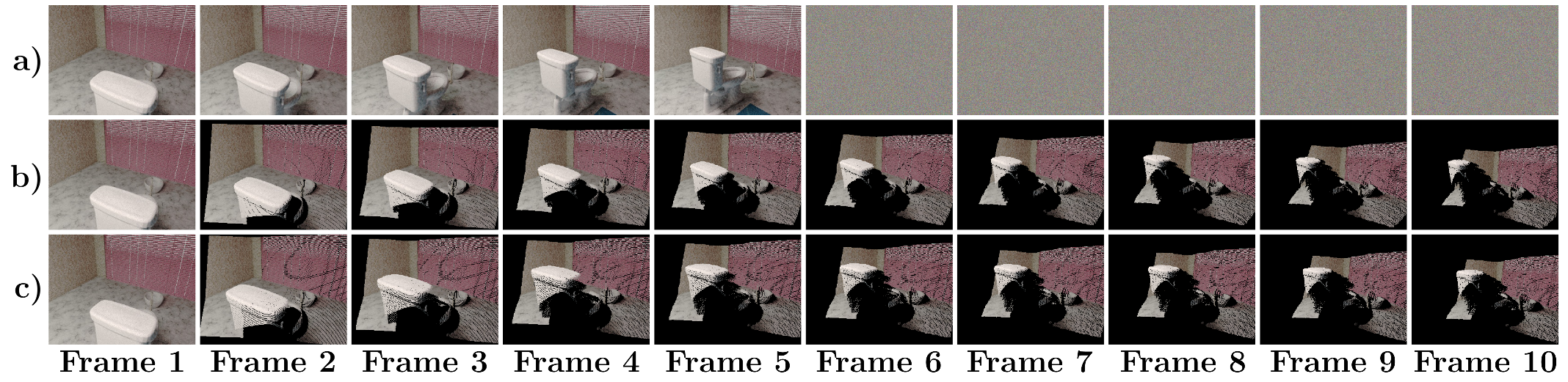}
		\vspace{-0.7cm}	
		\caption{Dynamic toy-data. a): Input sequence; b): Ground-truth projection of frame one; c): Geometric projection of frame one using motion estimates of FMTNet.}
		\label{fig:motion_integration}	
		\vspace{-0.1cm}
\end{figure}

\noindent\textbf{Temporal Motion Integration.} In order to obtain a meaningful motion estimate for images that do not contain any information, it is essential to propagate and aggregate dynamics over time. Using a dynamic toy-dataset, we evaluate the ability of our motion filter to perform these two tasks. The dataset contains sequences of length 10 for which we have replaced the last five frames with Gaussian noise. In Fig.~\ref{fig:motion_integration} and Tab.~\ref{table:motion_integration}, we report the performance of our motion filter, which has been fine-tuned on the dynamic toy-data. We use the translation norm~$\Delta T$ (Eq.~\ref{eq:motion_loss_trans}) and the rotation angle $\Delta R$ (Eq.~\ref{eq:motion_loss_rot}) of the relative transformation between predicted and gound-truth motion as evaluation metrics. 
\begin{table}[h]
			\vspace{-0.0cm}
			\begin{center}
			\scalebox{0.71}{
				\begin{tabular}[]{|c||c|c|c|c||c|c|c|c|c|}
					\hline
										& Frame 1-2 & Frame 2-3 & Frame 3-4	& Frame 4-5 & Frame 5-6 	
										& Frame 6-7 & Frame 7-8	& Frame 8-9	& Frame 9-10	  		\\
					\hline 
					$\Delta T$  		& $0.0073$ & $0.0047$ & $0.0042$ & $0.0041$ & $0.0069$ 
										& $0.0094$ & $0.0112$ & $0.0127$ & $0.0139$ \\	                

					$\Delta R$ 			& $0.0339$ & $0.0209$ & $0.0193$ & $0.0189$ & $0.0279$ 
										& $0.0303$ & $0.0317$ & $0.0322$ & $0.0336$ \\                 
					\hline			
				\end{tabular}		
			}		 
		\end{center}
	\vspace{-0.1cm}
	\caption{Translation norm $\Delta T$ and rotation angle $\Delta R$ of FMTNet on partially temporally occluded dynamic toy-data, computed for each frame pair using a test set of 30,000 sequences.}
	\label{table:motion_integration}
	\vspace{-0.4cm}
\end{table}

In the first four computation steps, the translation norm and rotation angle decrease, as the filter integrates information. In the next five steps, the filter still delivers meaningful predictions, which slowly get worse due to accumulating errors. In Fig.~\ref{fig:motion_integration}, we show the successive projection of the first frame, computed with ground-truth motions and predicted motions, respectively. We use the ground-truth depth maps for both successive projections.

\noindent\textbf{Comparison with baselines.}  To compare our model with the introduced baselines, we use the test set described in Section~\ref{sec:implementation_details}. In Tab.~\ref{table:final_results}, we report the Mean IoU of all models on a per-frame basis. Results show clear superiority of the filtered models (MFF-PSPNet, FMTNet), compared to the unfiltered baseline (MPSPNet). Only for the first frame, MPSPNet outperforms the filtered architectures. This is most likely due to not yet well-initialized hidden filter states. Our model surpasses the other filtered baseline and does not seem to be so strongly affected by poorly initialized hidden states. Unexpectedly, the performance of our model decreases again from Frame 5 forward. We suspect that this is due to the fairely simple design of our feature update module. A more sophisticated fusion approach could counter this behavior. We plan to further investigate this deficiency in the future. An example prediction of FMTNet is included in the supplementary material.
\begin{table}[h]
	\vspace{-0.1cm}
	\begin{center}
		\scalebox{0.92}{
			\begin{tabular}[]{|c||c|c|c|c|c|c|c|}
				\hline
								& Frame 1 & Frame 2 & Frame 3 & Frame 4 & Frame 5 
								& Frame 6 & Frame 7  		\\
				\hline 
				MPSPNet 		& $\mathbf{39.0\,\%}$ & $38.5\,\%$ & $37.9\,\%$ & $38.8\,\%$ & $38.4\,\%$
								& $38.1\,\%$ & $38.1\,\%$  	\\	            
				\hline 
				MFF-PSPNet 		& $37.1\,\%$ & $39.3\,\%$ & $39.9\,\%$ & $40.6\,\%$ & $40.2\,\%$
								& $40.3\,\%$ & $\mathbf{40.4\,\%}$ 	\\        
				\hline 
				FMTNet (ours)			& $38.9\,\%$ & $\mathbf{40.8\,\%}$ & $\mathbf{41.1\,\%}$ & $\mathbf{41.3\,\%}$ & $\mathbf{41.0\,\%}$
								& $\mathbf{40.9\,\%}$ & $\mathbf{40.4\,\%}$ \\	           									 
				\hline			
			\end{tabular}		             
		}		 
	\end{center}
	\vspace{-0.2cm}
	\caption{Mean IoU of all models on test sequences which are perturbed by aleatoric noise. MPSPNet: unfiltered baseline; MFF-PSPNet: filtered baseline; FMTNet: our model.}
	\label{table:final_results}
	\vspace{-0.5cm}
\end{table}

\section{Conclusion}
\label{sec:conclusion}
In this paper, we have introduced a functionally modularized temporal representation filter to tackle aleatoric failures of a single frame segmentation model. The main idea behind the filter is to decompose the filter task into less complex and more transparent subtasks. The resulting filter consists of multiple submodules, which can be pre-trained, debugged, and evaluated independently. In contrast to many other approaches in the literature, our filter also works in challenging situation, \eg brief sensor outages. Using a simulated dataset, we showed the superiority of our model compared to classical baselines. In the future, we plan to extend our filter to explicitly model dynamic objects in the scene. 

\bibliography{bmvc18}

\begin{thebibliography}{28}
\providecommand{\natexlab}[1]{#1}
\providecommand{\url}[1]{\texttt{#1}}
\expandafter\ifx\csname urlstyle\endcsname\relax
  \providecommand{\doi}[1]{doi: #1}\else
  \providecommand{\doi}{doi: \begingroup \urlstyle{rm}\Url}\fi

\bibitem[Cho et~al.(2014)Cho, Van~Merri{\"e}nboer, Gulcehre, Bahdanau,
  Bougares, Schwenk, and Bengio]{cho_2014_arxiv}
Kyunghyun Cho, Bart Van~Merri{\"e}nboer, Caglar Gulcehre, Dzmitry Bahdanau,
  Fethi Bougares, Holger Schwenk, and Yoshua Bengio.
\newblock Learning phrase representations using {RNN} encoder-decoder for
  statistical machine translation.
\newblock \emph{Preprint arXiv:1406.1078}, 2014.

\bibitem[Cordts et~al.(2016)Cordts, Omran, Ramos, Rehfeld, Enzweiler, Benenson,
  Franke, Roth, and Schiele]{cordts_2016_CVPR}
Marius Cordts, Mohamed Omran, Sebastian Ramos, Timo Rehfeld, Markus Enzweiler,
  Rodrigo Benenson, Uwe Franke, Stefan Roth, and Bernt Schiele.
\newblock The {C}ityscapes dataset for semantic urban scene understanding.
\newblock In \emph{IEEE Conference on Computer Vision and Pattern Recognition
  (CVPR)}, 2016.

\bibitem[Doll\'ar et~al.(2009)Doll\'ar, Wojek, Schiele, and
  Perona]{dollar_2019_CVPR}
P.~Doll\'ar, C.~Wojek, B.~Schiele, and P.~Perona.
\newblock Pedestrian detection: A benchmark.
\newblock In \emph{IEEE Conference on Computer Vision and Pattern Recognition
  (CVPR)}, 2009.

\bibitem[Fayyaz et~al.(2016)Fayyaz, Saffar, Sabokrou, Fathy, Klette, and
  Huang]{fayyaz_2016_arxiv}
Mohsen Fayyaz, Mohammad~Hajizadeh Saffar, Mohammad Sabokrou, Mahmood Fathy,
  Reinhard Klette, and Fay Huang.
\newblock {STFCN}: Spatio-temporal {FCN} for semantic video segmentation.
\newblock \emph{Preprint arXiv:1608.05971}, 2016.

\bibitem[Gadde et~al.(2017)Gadde, Jampani, and Gehler]{gadde_2017_ICCV}
Raghudeep Gadde, Varun Jampani, and Peter~V. Gehler.
\newblock Semantic video {CNNs} through representation warping.
\newblock In \emph{IEEE International Conference on Computer Vision (ICCV)},
  2017.

\bibitem[Jin et~al.(2016)Jin, Li, Xiao, Shen, Lin, Yang, Chen, Dong, Liu, Jie,
  et~al.]{jin_2016_arxiv}
Xiaojie Jin, Xin Li, Huaxin Xiao, Xiaohui Shen, Zhe Lin, Jimei Yang, Yunpeng
  Chen, Jian Dong, Luoqi Liu, Zequn Jie, et~al.
\newblock Video scene parsing with predictive feature learning.
\newblock \emph{Preprint arXiv:1612.00119}, 2016.

\bibitem[Kendall and Gal(2017)]{kendall_2017_arxiv}
Alex Kendall and Yarin Gal.
\newblock What uncertainties do we need in {B}ayesian deep learning for
  computer vision?
\newblock \emph{Preprint arXiv:1703.04977}, 2017.

\bibitem[Kingma and Ba(2014)]{kingma_2014_arxiv}
Diederik~P Kingma and Jimmy Ba.
\newblock Adam: A method for stochastic optimization.
\newblock \emph{Preprint arXiv:1412.6980}, 2014.

\bibitem[Kiureghian and Ditlevsen(2009)]{Kiureghian_2019_Safety}
Armen~Der Kiureghian and Ove Ditlevsen.
\newblock Aleatory or epistemic? {D}oes it matter?
\newblock \emph{Structural Safety}, 31\penalty0 (2):\penalty0 105 -- 112, 2009.
\newblock Risk Acceptance and Risk Communication.

\bibitem[Kundu et~al.(2016)Kundu, Vineet, and Koltun]{kundu_2016_CVPR}
Abhijit Kundu, Vibhav Vineet, and Vladlen Koltun.
\newblock Feature space optimization for semantic video segmentation.
\newblock In \emph{IEEE Conference on Computer Vision and Pattern Recognition
  (CVPR)}, pages 3168--3175, 2016.

\bibitem[Lei and Todorovic(2016)]{lei_2016_ECCV}
Peng Lei and Sinisa Todorovic.
\newblock Recurrent temporal deep field for semantic video labeling.
\newblock In \emph{European Conference on Computer Vision (ECCV)}, pages
  302--317, 2016.

\bibitem[Mahjourian et~al.(2016)Mahjourian, Wicke, and
  Angelova]{mahjourian_2016_arxiv}
Reza Mahjourian, Martin Wicke, and Anelia Angelova.
\newblock Geometry-based next frame prediction from monocular video.
\newblock \emph{Preprint arXiv:1609.06377}, 2016.

\bibitem[Mahjourian et~al.(2018)Mahjourian, Wicke, and
  Angelova]{mahjourian_2018_arxiv}
Reza Mahjourian, Martin Wicke, and Anelia Angelova.
\newblock Unsupervised learning of depth and ego-motion from monocular video
  using {3D} geometric constraints.
\newblock \emph{Preprint arXiv:1802.05522}, 2018.

\bibitem[McCormac et~al.(2017)McCormac, Handa, Leutenegger, and
  Davison]{mccormac_2017_ICCV}
John McCormac, Ankur Handa, Stefan Leutenegger, and Andrew~J Davison.
\newblock {SceneNet} {RGB-D}: Can {5M} synthetic images beat generic {ImageNet}
  pre-training on indoor segmentation.
\newblock In \emph{IEEE International Conference on Computer Vision (ICCV)},
  2017.

\bibitem[Nilsson and Sminchisescu(2016)]{Nilsson_2016_CoRR}
David Nilsson and Cristian Sminchisescu.
\newblock Semantic video segmentation by gated recurrent flow propagation.
\newblock \emph{CoRR}, abs/1612.08871, 2016.

\bibitem[Pavel et~al.(2017)Pavel, Schulz, and Behnke]{pavel_2015_IJCNN}
Mircea~Serban Pavel, Hannes Schulz, and Sven Behnke.
\newblock Object class segmentation of {RGB-D} video using recurrent
  convolutional neural networks.
\newblock \emph{Neural Networks, Elsevier}, 2017.

\bibitem[Radwan et~al.(2018)Radwan, Valada, and Burgard]{radwan_2018_arxiv}
Noha Radwan, Abhinav Valada, and Wolfram Burgard.
\newblock {VLocNet++}: Deep multitask learning for semantic visual localization
  and odometry.
\newblock \emph{Preprint arXiv:1804.08366}, 2018.

\bibitem[Ruder(2017)]{ruder_2017_arxiv}
Sebastian Ruder.
\newblock An overview of multi-task learning in deep neural networks.
\newblock \emph{Preprint arXiv:1706.05098}, 2017.

\bibitem[Tran et~al.(2016)Tran, Bourdev, Fergus, Torresani, and
  Paluri]{tran_2016_CVPR}
Du~Tran, Lubomir Bourdev, Rob Fergus, Lorenzo Torresani, and Manohar Paluri.
\newblock Deep end2end voxel2voxel prediction.
\newblock In \emph{IEEE Conference on Computer Vision and Pattern Recognition
  Workshops}, pages 17--24, 2016.

\bibitem[Ummenhofer et~al.(2017)Ummenhofer, Zhou, Uhrig, Mayer, Ilg,
  Dosovitskiy, and Brox]{ummenhofer_2017_CVPR}
Benjamin Ummenhofer, Huizhong Zhou, Jonas Uhrig, Nikolaus Mayer, Eddy Ilg,
  Alexey Dosovitskiy, and Thomas Brox.
\newblock {DeMoN}: Depth and motion network for learning monocular stereo.
\newblock In \emph{IEEE Conference on Computer Vision and Pattern Recognition
  (CVPR)}, 2017.

\bibitem[Valipour et~al.(2017)Valipour, Siam, Jagersand, and
  Ray]{valipour_2017_WACV}
Sepehr Valipour, Mennatullah Siam, Martin Jagersand, and Nilanjan Ray.
\newblock Recurrent fully convolutional networks for video segmentation.
\newblock In \emph{IEEE Winter Conference on Applications of Computer Vision
  (WACV)}, pages 29--36, 2017.

\bibitem[Vijayanarasimhan et~al.(2017)Vijayanarasimhan, Ricco, Schmid,
  Sukthankar, and Fragkiadaki]{Vijayanarasimhan_2017_corr}
Sudheendra Vijayanarasimhan, Susanna Ricco, Cordelia Schmid, Rahul Sukthankar,
  and Katerina Fragkiadaki.
\newblock {SfM-Net}: Learning of structure and motion from video.
\newblock \emph{CoRR}, abs/1704.07804, 2017.

\bibitem[Vu et~al.(2018)Vu, Choi, Schulter, and Chandraker]{vu_2018_arxiv}
Tuan-Hung Vu, Wongun Choi, Samuel Schulter, and Manmohan Chandraker.
\newblock Memory warps for learning long-term online video representations.
\newblock \emph{Preprint arXiv:1803.10861}, 2018.

\bibitem[Wagner et~al.(2018)Wagner, Fischer, Herman, and
  Behnke]{wagner_2018_ESANN}
J{\"o}rg Wagner, Volker Fischer, Michael Herman, and Sven Behnke.
\newblock Hierarchical recurrent filtering for fully convolutional densenets.
\newblock In \emph{European Symposium on Artificial Neural Networks (ESANN)},
  2018.

\bibitem[Yin and Shi(2018)]{yin_2018_arxiv}
Zhichao Yin and Jianping Shi.
\newblock {GeoNet}: Unsupervised learning of dense depth, optical flow and
  camera pose.
\newblock \emph{Preprint arXiv:1803.02276}, 2018.

\bibitem[Zhang et~al.(2014)Zhang, Jiang, Zhang, Li, Xia, and
  Chen]{zhang_2014_ICVRV}
Han Zhang, Kai Jiang, Yu~Zhang, Qing Li, Changqun Xia, and Xiaowu Chen.
\newblock Discriminative feature learning for video semantic segmentation.
\newblock In \emph{International Conference on Virtual Reality and
  Visualization (ICVRV)}, pages 321--326, 2014.

\bibitem[Zhao et~al.(2017)Zhao, Shi, Qi, Wang, and Jia]{zhao_2017_CVPR}
Hengshuang Zhao, Jianping Shi, Xiaojuan Qi, Xiaogang Wang, and Jiaya Jia.
\newblock Pyramid scene parsing network.
\newblock In \emph{IEEE Conf. on Computer Vision and Pattern Recognition
  (CVPR)}, pages 2881--2890, 2017.

\bibitem[Zhou et~al.(2017)Zhou, Brown, Snavely, and Lowe]{zhou_2017_CVPR}
Tinghui Zhou, Matthew Brown, Noah Snavely, and David~G. Lowe.
\newblock Unsupervised learning of depth and ego-motion from video.
\newblock In \emph{IEEE Conference on Computer Vision and Pattern Recognition
  (CVPR)}, 2017.

\end{thebibliography}

\clearpage
\setcounter{section}{0}
\setcounter{table}{0}
\setcounter{figure}{0}
\setcounter{equation}{0}
\setcounter{page}{1}
\bmvaResetAuthors

\title{Supplementary Material - Functionally Modular and Interpretable Temporal Filtering for Robust Segmentation}

\addauthor{J\"org Wagner\textsuperscript{\scriptsize{1,}}}{Joerg.Wagner3@de.bosch.com}{2}
\addauthor{Volker Fischer}{Volker.Fischer@de.bosch.com}{1}
\addauthor{Michael Herman}{Michael.Herman@de.bosch.com}{1}
\addauthor{Sven Behnke}{behnke@cs.uni-bonn.de}{2}

\addinstitution{
	Bosch Center for Artificial Intelligence,\\
	71272 Renningen, Germany
}
\addinstitution{
	University of Bonn,\\
	Computer Science Institute VI,\\
	Autonomous Intelligent Systems,\\
	Endenicher Allee 19 A,\\
	53115 Bonn, Germany
}

\supp_title

\section{Simulation of Aleatoric Perturbations}
\label{sec:supp_noise}
\begin{figure}[h]
	\includegraphics[width=\linewidth]{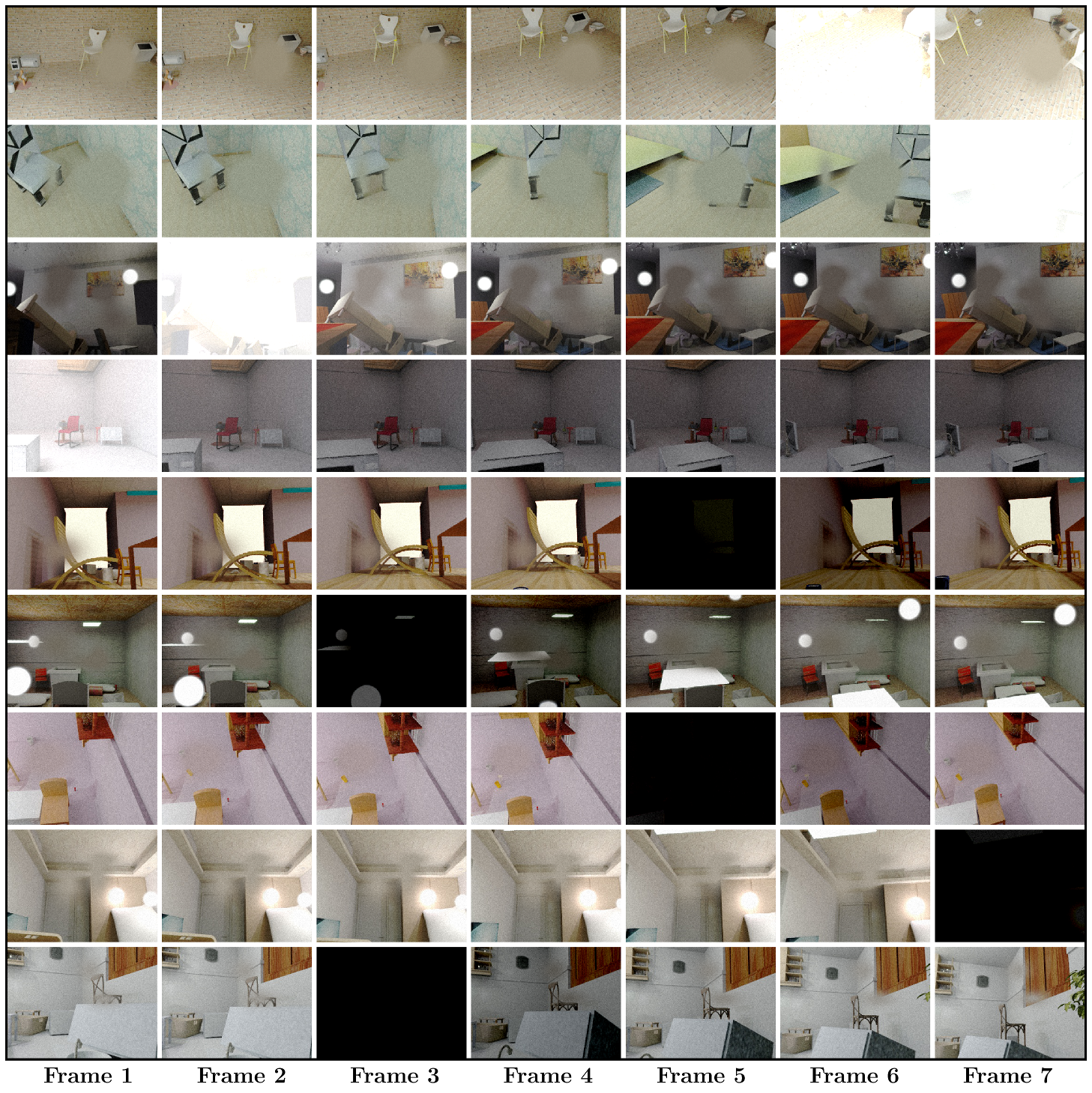}
	\vspace{-1.5em}
	\caption{Example sequences of the data used for training and in the evaluation. One sequence of length 7 is shown per row. Each sequence is perturbed with noise, clutter, and changes in lighting conditions.}
	\label{fig:example_sequences}
\end{figure}

Aleatoric failures originate from perturbations inherent in the data. To simulate such perturbations, we add noise, clutter, and changes in lighting conditions to all sequences. In the following, we give a detailed desciption of the process used to generate these perturbations. After applying the perturbations to the clean sequences generated from the \textit{SceneNet RGB-D} dataset, we clip pixels to the interval~$[0, 1]$ to get valid images. Example sequences are shown in Fig.~\ref{fig:example_sequences}. \\

\noindent\textbf{Noise} is simulated by adding independent Gaussian noise with zero mean to each pixel. The variance of the noise is independently sampled for each sequence from the interval~$[0, 0.001]$. \\

\noindent\textbf{Clutter} is introduced by setting subregions of each image to the pixel mean $\mathbf{\mu}$, computed on a per-sequence basis. The clutter is generated once per-sequence and applied to each frame. Thus, the resulting clutter pattern is the same in each frame, comparable to dirt on the camera lens. The perturbed images $\mathbf{x}_{j}\sp{\prime}$ are calculated by: 
\begin{equation}
\mathbf{x}_{j}\sp{\prime} = \mathbf{x}_{j} \cdot (1 - \mathbf{m}) + \mathbf{\mu} \cdot \mathbf{m},
\end{equation}
where $\mathbf{m}$ is a per-sequence clutter mask, $\mathbf{\mu}$ the per-sequence pixel mean, and $\mathbf{x}_{j}$ the clean image. The clutter mask is generated by summing $N_{g}$ Gaussian kernels whose centers are randomly placed (uniformly sampled) within the image dimensions. Each kernel is normalized to the maximum value one. The number of kernels $N_{g}$ is uniformly sampled for each sequence from the interval~$[0, 8]$. In addition, we uniformly sample the standard deviation of each dimension independently from the interval~$[10, 36]$. The kernels are truncated at three times the standard deviation.\\

\noindent\textbf{Changes in lighting conditions} are simulated by increasing or decreasing the intensity of frames. For each sequence, we uniformly sample one frame $\mathbf{x}_{i}$ and a scaling factor $s$ from the interval~$[0.5, 1.0]$. In addition, we draw a multiplier $p$ which with a probability of 0.5 is either 1 or -1. The perturbed images $\mathbf{x'}_{j}$ are calculated by:
\begin{align}
\forall j < i: \mathbf{x}_{j}\sp{\prime} &= \mathbf{x}_{j}, \\
\forall j \ge i: \mathbf{x}_{j}\sp{\prime} &= \mathbf{x}_{j} + p \cdot 0.3^{j-i} \cdot s.
\end{align}

\newpage
\section{Example Prediction of FMTNet}
\begin{figure}[h]
	\includegraphics[width=\linewidth]{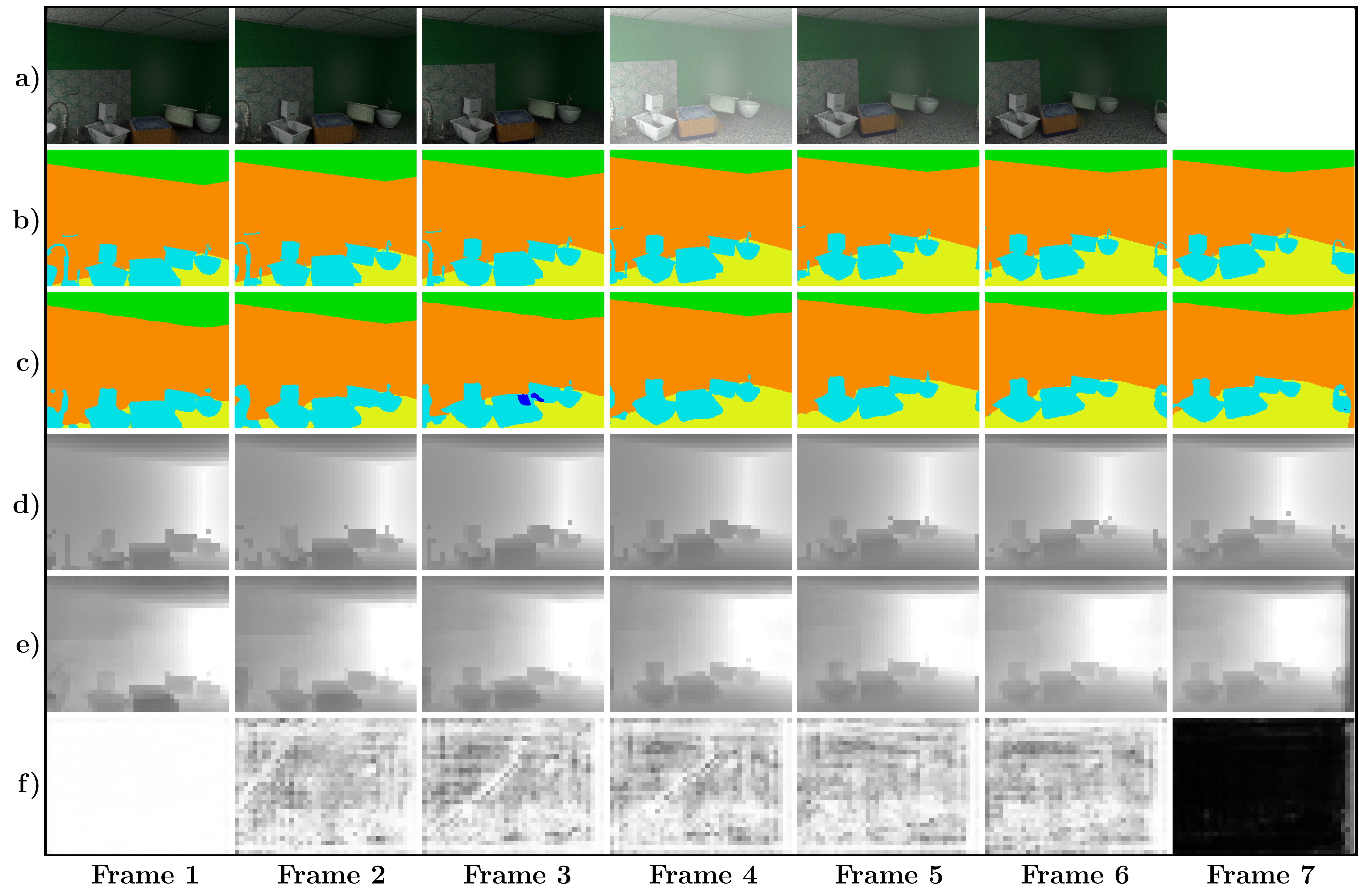}
	\vspace{-1.5em}
	\caption{Example prediction of our FMTNet. a): Input sequence; b): Ground-truth semantic segmentation. c): Predicted semantic segmentation; d): Ground-truth deph; e): Predicted depth; f): Update gate $\mathbf{i}_{t}$, white corresponds to a value of one (use new information).}
	\label{fig:fmtnet_pred}
\end{figure}

In Fig.~\ref{fig:fmtnet_pred}, we show an example prediction of our FMTNet. In addition to visualizing the predicted semantic segmentation (Fig.~\ref{fig:fmtnet_pred}\textcolor{red}{c}), we also show the predicted depth map (Fig.~\ref{fig:fmtnet_pred}\textcolor{red}{e}) and the update gate (Fig.~\ref{fig:fmtnet_pred}\textcolor{red}{f}), which are two of the human interpretable representations computed within our functionally modularized temporal filter. The model is able to predict a meaningful depth map as well as camera motion, which are required to propagate information over time. This is especially visible in the last frame of the sequence -- although the last frame is missing, the model is still able to produce a meaningful semantic segmentation. In Fig.~\ref{fig:fmtnet_pred}\textcolor{red}{f}, we show the gate $\mathbf{i}_{t}$ of our update module. A white pixel corresponds to a gate value of one, which means that the model uses information provided by the current input frame. A black pixel, on the other hand, corresponds to a gate value of zero -- the model relies on prior knowledge of previous frames. As expected, the gate of the first frame is fully white, since the filter has to rely on new information. In the last frame, the gate is mainly black, since no meaningful information is provided in that frame. The gate values at the right border of all frames are more white, as the model has never seen these areas before due to camera motion. 

\end{document}